\documentclass{article}

\usepackage[main, final]{neurips_2026}

\usepackage[utf8]{inputenc}
\usepackage[T1]{fontenc}
\usepackage{hyperref}
\usepackage{url}
\usepackage{booktabs}
\usepackage{amsfonts}
\usepackage{amsmath}
\usepackage{amssymb}
\usepackage{amsthm}
\usepackage{nicefrac}
\usepackage{microtype}
\usepackage{xcolor}
\usepackage{graphicx}
\usepackage{multirow}
\usepackage{colortbl}
\usepackage{caption}
\usepackage{placeins}
\usepackage{algorithm}
\usepackage{tabularx}
\usepackage{algpseudocode}
\usepackage{tikz}
\usetikzlibrary{arrows.meta, positioning, calc}

\newtheorem{proposition}{Proposition}
\definecolor{myrow}{RGB}{220,220,220}

\title{IdealCache: Rethinking Cache Scheduling in Diffusion Transformers via Ideal Trajectories}

\author{%
  Shang Xue$^{1}$\thanks{Equal contribution.} \quad
  Li Zhu$^{2}$\footnotemark[1] \quad
  Ping Chen$^{3}$\thanks{Corresponding author.} \\
  $^{1}$Independent Researcher, Santa Clara, California, United States \\
  $^{2}$Independent Researcher, Mountain View, California, United States \\
  $^{3}$University of Electronic Science and Technology of China \\
  {\small\texttt{\{shangxue.sdu,redcping\}@gmail.com, lizhu2@alumni.cmu.edu}}
}

\begin{document}

\maketitle

\begin{abstract}
Feature caching accelerates Diffusion Transformers by reusing intermediate
computations across denoising timesteps, yet existing methods share a
fundamental blind spot: they treat the standard reference denoising
trajectory as the optimisation target, approximating its per-step states
and thus inheriting its output as an implicit quality ceiling. We argue
that this target is arbitrary --- what truly matters is the \emph{ideal
result}, not the intermediate path taken to reach it. We propose
\textbf{IdealCache}, a training-free framework that abandons step-wise
imitation of the reference trajectory. Instead, it anchors caching on a
per-input ensemble of \emph{ideal trajectories} constructed from
candidate target results, aggregates their geometry into a single
\emph{graph}, and discovers optimal non-uniform schedules via a
constrained shortest path on this graph. By decoupling from the
reference process, IdealCache breaks the performance bottleneck of
trajectory-approximation; our schedules can attain near-lossless
fidelity in generation and even surpass on task-specific metrics
(e.g., maintaining higher semantic consistency in image editing).
Without retraining, IdealCache achieves $6.35\times$ speedup on
Qwen-Image, $5.84\times$ on Qwen-Image-Edit, and $4.32\times$ on
HunyuanVideo, with competitive or superior quality across tasks.
\end{abstract}

\begin{figure}[h]
\centering
\includegraphics[width=1\linewidth]{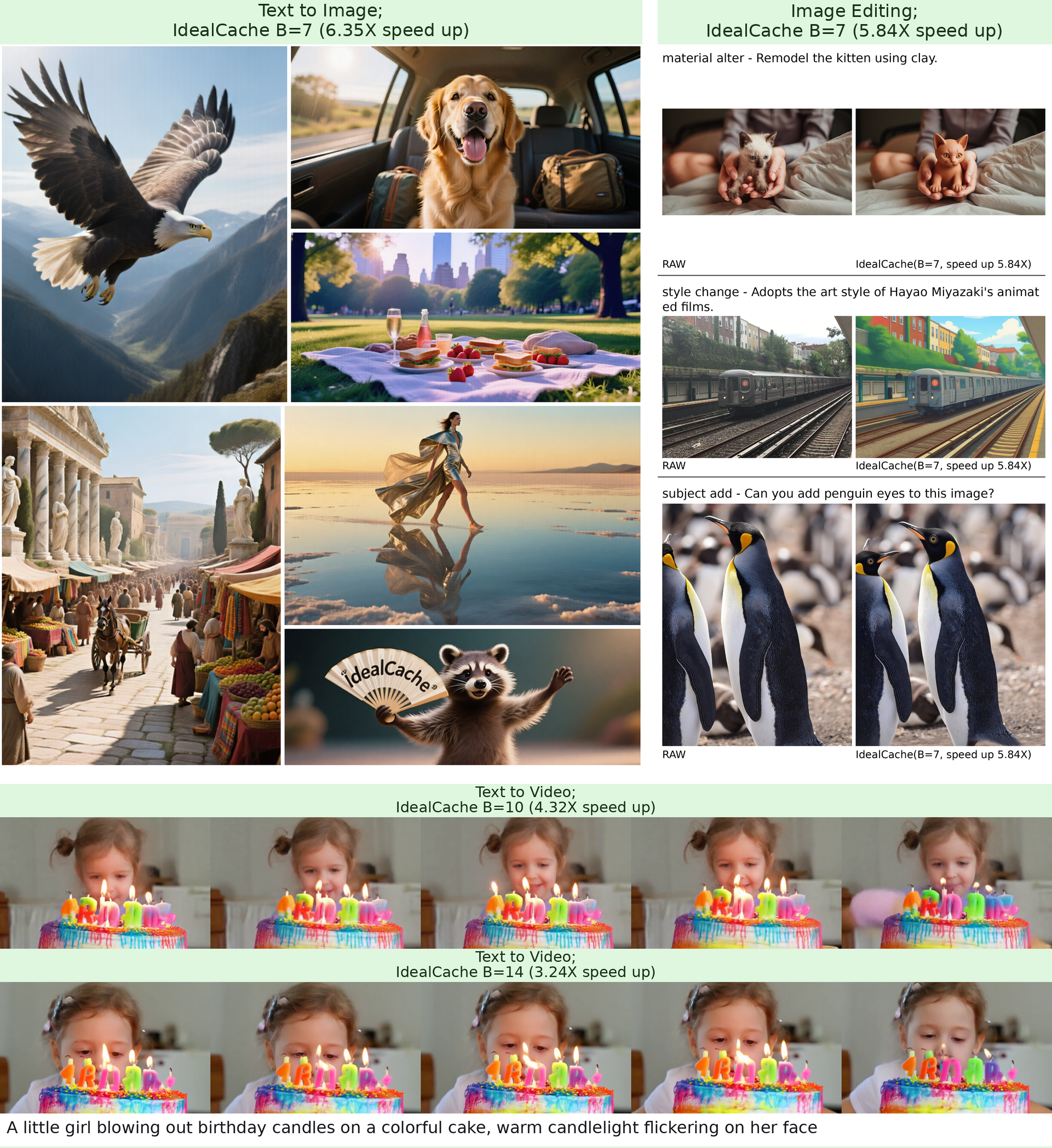}
\caption{Representative IdealCache outputs across diverse prompts and
tasks (Qwen-Image, Qwen-Image-Edit, HunyuanVideo1.0). Without retraining, IdealCache delivers high-fidelity generations.}
\label{fig:teaser}
\end{figure}

%%%%%%%%%%%%%%%%%%%%%%%%%%%%%%%%%%%%%%%%%%%%%%%%%%%%%%%%%%%%
\section{Introduction}
%%%%%%%%%%%%%%%%%%%%%%%%%%%%%%%%%%%%%%%%%%%%%%%%%%%%%%%%%%%%

Diffusion Transformers (DiTs)~\citep{peebles2023dit} now drive
state-of-the-art image and video generation, but their iterative sampling
remains a deployment bottleneck: each output requires many transformer
forward passes, each of which is expensive at the resolutions and model
scales used in practice. Feature caching is one of the most effective
training-free responses to this cost. By reusing residual computations
across consecutive denoising timesteps, caching skips a significant
fraction of full forward passes without retraining the
model~\citep{ma2024deepcache,selvaraju2024fora,zou2024toca,liu2025taylorseer}.
Methods in this line have steadily refined how to cache --- which
layers, which tokens, what extrapolation rule. But almost all of them
share the same answer to a more fundamental question: \emph{when}
should the cache be refreshed? The standard answer is to fix a
hand-designed schedule --- typically computed every $K$-th step over a
uniform $50$-step grid --- and judge the cached output by how closely
it tracks the unmodified $50$-step denoising trajectory, which we will
call the \emph{vanilla pipeline}. This paper argues that the choice of
\emph{when} matters as much as the choice of \emph{how}, and that the
schedule deserves to be optimised rather than hand-set.

\noindent \textbf{Rethinking what we are matching.}
Tying the cached schedule to the vanilla pipeline has a subtle but
consequential effect: the pipeline's output becomes the implicit
ceiling of the schedule, because the only way to look ``better'' under
a trajectory-tracking objective is to be the pipeline. We argue this is
the wrong ceiling. What the user actually wants is the \emph{ideal
result} for the input; the trajectory taken to reach that result is a
means, not an end. If we re-anchor the optimisation on the ideal result
rather than on the vanilla pipeline's intermediate states, the schedule
is freed to spend compute where it most helps the result --- and the
ceiling is no longer structurally bounded by the vanilla pipeline.

\noindent \textbf{Cache scheduling as graph search on the ideal trajectory.}
Concretely, given a candidate target result for the input we draw the
straight line in latent space from noise to that target --- the
\emph{ideal trajectory}. Every cached interval $k\to t$ has a
measurable cost on this trajectory: do one full pass at $t$ to populate
the cache, reuse those residuals to take $t-k$ Euler updates along the
way, and compare where you land to where the ideal trajectory says you
should be. This squared distance is the \emph{cache deviation}, and we
use it directly as the edge weight of a directed acyclic graph over
candidate timesteps. Averaging the edge weights across an ensemble of
candidate targets gives a single graph $\bar{\mathcal{G}}$, on
which the optimal length-$B$ schedule is the constrained shortest path
--- solved exactly in polynomial time. There is no learning, no tuning,
and no architectural change.

\noindent \textbf{What this buys us.}
Because the optimisation is anchored on the ideal trajectory rather
than on the vanilla pipeline, the resulting schedules are no longer
structurally bounded by the pipeline's output. Empirically, IdealCache
attains quality close to the $50$-step vanilla pipeline at much smaller
budgets across all three modalities we test --- text-to-image, image
editing, and text-to-video --- and on certain task-specific metrics
(most notably semantic consistency in image editing) the optimised
schedules can match and modestly exceed the vanilla pipeline. We
deliberately avoid claiming universal superiority over the $50$-step
output: the vanilla pipeline is still the implementation anchor for our
reference target results, and on aggregate quality metrics such as
ImageReward we remain below the $50$-step output at higher compression
ratios. We do claim that treating the schedule as a search problem on
the ideal trajectory yields a strictly better quality--budget frontier
than the hand-designed schedules used by prior caching methods.

Our contributions are:
% \vspace{-7pt}
\begin{itemize}\setlength\itemsep{0em}
    \item \textbf{Cache scheduling as ideal-trajectory geometry.} We
    re-frame cache scheduling as minimum-deviation path selection
    against a per-input ideal trajectory, removing the dependence of
    cached schedules on the vanilla pipeline's intermediate states.
    \item \textbf{Cache deviation as a tractable edge weight.} We show
    that the deviation of a cached rollout from the ideal trajectory
    can be computed directly from a single full pass and used, without
    learning, as the edge weight of an exact shortest-path search.
    \item \textbf{Pareto-dominant quality--budget frontier.} On
    Qwen-Image, Qwen-Image-Edit, and HunyuanVideo, the schedules found
    by IdealCache Pareto-dominate prior caching baselines at every
    speed group we evaluate, and on image editing they reach quality
    competitive with --- and on certain metrics modestly above --- the
    $50$-step vanilla pipeline, as shown in Figure~\ref{fig:teaser}.
\end{itemize}

%%%%%%%%%%%%%%%%%%%%%%%%%%%%%%%%%%%%%%%%%%%%%%%%%%%%%%%%%%%%
% \vspace{-10pt}
\section{Related Work}
%%%%%%%%%%%%%%%%%%%%%%%%%%%%%%%%%%%%%%%%%%%%%%%%%%%%%%%%%%%%

\noindent \textbf{Diffusion model acceleration.}
Accelerating diffusion models generally involves improving ODE solver efficiency or reducing neural function evaluations. Methods like DDIM~\citep{song2021ddim}, DPM-Solver and DPM-Solver++~\citep{lu2022dpm, lu2022dpmpp}, DEIS~\citep{zhang2023deis}, and UniPC~\citep{zhao2023unipc} design advanced integrators tailored to the diffusion probability flow. Alternatively, Consistency Models~\citep{song2023consistency}, distillation approaches~\citep{salimans2022distillation, luo2023lcm}, and flow-straightening techniques~\citep{liu2023flow, liu2024instaflow} train models to generate high-quality outputs in just a few steps. While highly effective, these methods primarily target the numerical integration problem; they are complementary to caching and do not explicitly optimise the schedule of expensive transformer evaluations. Orthogonal to the temporal axis, Just-in-Time (JiT)~\citep{sun2026jit} accelerates DiTs in the spatial domain, evolving the full latent from computations on a dynamically selected sparse subset of anchor tokens.

\noindent \textbf{Feature caching for diffusion transformers.}
Caching exploits temporal smoothness to skip redundant computations by reusing residuals across steps. DeepCache~\citep{ma2024deepcache} introduced this in U-Nets, and subsequent works adapted it to DiTs via explicit residual reuse ($\Delta$-DiT~\citep{chen2024deltadit}, FORA~\citep{selvaraju2024fora}), token-level selection (ToCa~\citep{zou2024toca}, DuCa~\citep{zou2024duca}), and attention broadcasting (PAB~\citep{zhao2024pab}). Others employ predictors to estimate feature changes, using timestep embeddings (TeaCache~\citep{liu2024teacache}), Taylor expansions (TaylorSeer~\citep{liu2025taylorseer}), inferred ODE mixtures (HyCa~\citep{zheng2025hyca}), or trainable layer selectors (Learning-to-Cache~\citep{ma2024ltc}). More recent work verifies speculatively forecast features before accepting them (SpeCa~\citep{liu2025speca}), compensates caching errors with cached gradients (GOC~\citep{qiu2025goc}), or calibrates an offline cache policy together with timestep adjustment and residual error rectification (ERTACache~\citep{peng2026ertacache}). Crucially, these methods share a fundamental limitation: their caching intervals are fixed and input-agnostic. Furthermore, they evaluate cached paths by how well they approximate the reference denoising trajectory, implicitly inheriting the reference output as a strict quality ceiling. IdealCache breaks both of these assumptions.

\noindent \textbf{Adaptive caching and timestep scheduling.}
Independent of caching, timestep scheduling has been explored to minimise ODE truncation errors (EDM~\citep{karras2022elucidating}), fit trajectory regularity (GITS~\citep{chen2024trajectory}), or bound KL divergence (Align Your Steps~\citep{sabour2024align}). Concurrently, content-aware caching methods like AdaCache~\citep{kahatapitiya2024adacache}, FasterCache~\citep{lv2025fastercache}, and DiCache~\citep{bu2025dicache} introduce runtime heuristics to dynamically skip steps based on local feature or signal differences. ShortDF~\citep{chen2025shortdf} casts the denoising process as a shortest-path problem over timesteps to minimise reconstruction residuals. Recently, \emph{Chain-of-Trajectories} (CoTj)~\citep{chen2026cotj} advanced scheduling by casting it as a content-adaptive graph-planning problem. IdealCache unifies these directions. Building upon CoTj's shortest-path machinery, we formulate cache scheduling as a globally optimal graph search. Crucially, rather than relying on local greedy heuristics or tracking the vanilla pipeline's intermediate states, we anchor our search on a per-input \emph{ideal} trajectory, utilizing cache deviations as edge weights to directly recover the optimal schedule under a fixed step budget.

\section{Method}
%%%%%%%%%%%%%%%%%%%%%%%%%%%%%%%%%%%%%%%%%%%%%%%%%%%%%%%%%%%%

% This section describes IdealCache. We first review the rectified flow
% denoising process and explain how feature caching skips computation
% (Sec.~\ref{sec:prelim}). We then describe the core idea --- anchoring the
% optimisation on a per-input ensemble of \emph{ideal trajectories}
% constructed from candidate target results, rather than imitating the
% reference denoising path (Sec.~\ref{sec:ideal}) --- and finally show how
% to aggregate this ensemble into a graph and solve it for an optimal
% cache schedule (Sec.~\ref{sec:graph}). The full procedure is summarised
% in Algorithm~\ref{alg:main}.

\subsection{Preliminary}
\label{sec:prelim}

\noindent \textbf{Rectified flow denoising.}
We build on the rectified flow framework~\citep{liu2023flow}. Given a clean
image latent $x_0$ and Gaussian noise $\epsilon \sim \mathcal{N}(0,\mathbf{I})$,
the forward process defines a straight-line interpolation between data and
noise:
\begin{equation}
    x_t = (1-t)\,x_0 + t\,\epsilon, \quad t \in [0,1].
    \label{eq:rectflow}
\end{equation}
Here $t=0$ corresponds to the clean image and $t=1$ to pure noise. A
velocity network $v_\theta(x_t, t, \mathbf{c})$ is trained to predict the
direction from noise back to data, conditioned on $\mathbf{c}$ (e.g.\ a
prompt). At inference time we discretise the reverse process into $M$ first
order Euler steps:
\begin{equation}
    x_{t-\Delta t} = x_t - \Delta t \cdot v_\theta(x_t, t, \mathbf{c}).
    \label{eq:euler}
\end{equation}
Each step requires a full transformer forward pass, so the wall-clock cost
scales with $M$.

\noindent \textbf{Why caching works.} Between two consecutive timesteps, the input latent $x_t$ barely moves, so
the hidden states inside the transformer also change very little. This
means the \emph{residual} that each block adds (its output minus its input)
is nearly the same at $t$ and at $t-\Delta t$. Caching exploits exactly
this: at a step we keep, we run the full transformer and record the
residuals; at a step we skip, we reuse those recorded residuals instead of
running the transformer again.

Concretely, let $\mathbf{h}_t$ and $\mathbf{e}_t$ denote the image and text
hidden states after all $L$ transformer blocks at timestep $t$, and let
$\mathbf{h}_t^{\mathrm{in}}, \mathbf{e}_t^{\mathrm{in}}$ be the corresponding
inputs to the first block. We define the accumulated residuals
\begin{equation}
    \Delta\mathbf{h}_t = \mathbf{h}_t - \mathbf{h}_t^{\mathrm{in}},
    \qquad
    \Delta\mathbf{e}_t = \mathbf{e}_t - \mathbf{e}_t^{\mathrm{in}}.
    \label{eq:residuals}
\end{equation}
At a skipped timestep $t'$ we approximate the output of the transformer by
adding the cached residuals from a recent full step $t$:
\begin{equation}
    \mathbf{h}_{t'} \approx \mathbf{h}_{t'}^{\mathrm{in}} + \Delta\mathbf{h}_t,
    \qquad
    \mathbf{e}_{t'} \approx \mathbf{e}_{t'}^{\mathrm{in}} + \Delta\mathbf{e}_t,
    \label{eq:cache}
\end{equation}
which gives a cheap velocity estimate
$\hat{v}_{\mathrm{cache}}(x_{t'}, t';\, \Delta\mathbf{h}_t, \Delta\mathbf{e}_t)$
and the cached Euler update
\begin{equation}
    x_{t'-\Delta t} = x_{t'} - \Delta t\cdot
    \hat{v}_{\mathrm{cache}}\!\bigl(x_{t'},\, t';\,
    \Delta\mathbf{h}_t,\, \Delta\mathbf{e}_t\bigr).
    \label{eq:cache_euler}
\end{equation}
Existing caching methods~\citep{ma2024deepcache,lv2025fastercache,zou2024toca,zou2024duca,liu2024teacache} apply this approximation on a fixed, hand-designed schedule, typically a uniform 50-step schedule treated as the vanilla baseline, and evaluate the cached path against the reference denoising trajectory, which makes the reference output an implicit quality ceiling. The central question of this paper is: \emph{given a target result we actually want, which subset of steps should we compute to reach it under a fixed budget?}

\subsection{Ideal trajectory and cache deviation}
\label{sec:ideal}

\noindent \textbf{From imitating a process to anchoring on a result.}
A standard cached schedule, typically a uniform 50-step schedule, is
judged by how closely the cached states $\hat{x}_t$ track the states
$x_t$ of a reference denoising run. This makes the reference
\emph{process} the optimisation target, and the reference output is
inherited as a quality ceiling. We change the target. Suppose we
already had the ideal output $x_0^*$ for the input. The forward
interpolation in Eq.~\ref{eq:rectflow} then defines a straight line in
latent space:
\begin{equation}
    x_{t}^{*} = (1 - t)\,x_0^{*} + t\,\epsilon, \qquad t \in [0,1].
    \label{eq:ideal_traj}
\end{equation}
We call this line the \emph{ideal trajectory}: the locus of states a
sampler would pass through if it could go straight to $x_0^*$. The
ceiling becomes the ideal result itself, not any reference run's output.
We stress that this line is a \emph{target-anchored reference
trajectory}: it serves purely as a measurement anchor, and we do not
assume that the model's reverse dynamics follow it
(Proposition~\ref{prop:marginal}).

\noindent \textbf{Cache deviation along the ideal trajectory.}
With the ideal trajectory in hand, we now define the central object of
our method.
Suppose the sampler is currently at step $t$, and we wish to skip directly to a closer-to-data step $k < t$ without running the transformer at any intermediate step. To achieve this, we perform a single full forward pass at $t$ to populate the cache with residuals $(\Delta\mathbf{h}_t, \Delta\mathbf{e}_t)$, and then reuse these residuals to carry out $t-k$ Euler updates via Eq.~\ref{eq:cache_euler}:
\begin{equation}
    \hat{x}_{j-\Delta j} = \hat{x}_{j} - \Delta j\cdot
    \hat{v}_{\mathrm{cache}}\!\bigl(\hat{x}_{j},\, j;\,
    \Delta\mathbf{h}_t,\, \Delta\mathbf{e}_t\bigr),
    \quad \text{from } j=t \text{ down to } j=k,
    \label{eq:cache_rollout_short}
\end{equation}
where the rollout is initialized on the reference trajectory, $\hat{x}_t = x_t^{*}$, so that the measurement isolates the deviation introduced by reusing the residuals cached at $t$ over the interval $[k,t]$.

\begin{figure}[h]
\centering
\begin{tikzpicture}[>=Latex, scale=1.0, font=\small]
  % horizontal axis
  \draw[->, gray!70] (-0.2,0) -- (10.2,0);
  \node[gray!70, anchor=north] at (0.0,-0.05) {$t{=}1$};
  \node[gray!70, anchor=north] at (10.0,-0.05) {$t{=}0$};
  \node[gray!70, font=\footnotesize] at (5.0,-0.6) {denoising direction $\longrightarrow$};

  % ideal trajectory (a smooth curve, blue, descending right)
  \draw[blue!70!black, thick]
    plot[smooth, domain=0:10, samples=60] (\x, {3.0 - 0.20*\x - 0.018*\x*\x});
  \node[blue!70!black, anchor=south west] at (0.5,2.95) {ideal trajectory $x_t^*$};

  % node at t (where cache is captured)
  \coordinate (T) at (2.5, {3.0 - 0.20*2.5 - 0.018*2.5*2.5});
  \fill[blue!70!black] (T) circle (2.5pt);
  \node[blue!70!black, anchor=south] at (T) {$x_t^*$};

  % node at k on ideal (target)
  \coordinate (K) at (7.0, {3.0 - 0.20*7.0 - 0.018*7.0*7.0});
  \fill[blue!70!black] (K) circle (2.5pt);
  \node[blue!70!black, anchor=north west] at ([yshift=-1pt]K) {$x_k^*$};

  % cached rollout: dashed red, starts at T, ends offset above K
  \coordinate (Khat) at (7.0, {3.0 - 0.20*7.0 - 0.018*7.0*7.0 + 0.85});
  \draw[red!75!black, thick, dashed]
    plot[smooth, samples=40, domain=2.5:7.0]
    (\x, {(3.0 - 0.20*2.5 - 0.018*2.5*2.5)
          - 0.05*(\x-2.5)
          - 0.012*(\x-2.5)*(\x-2.5)});
  \fill[red!75!black] (Khat) circle (2.5pt);
  \node[red!75!black, anchor=south] at (Khat) {$\hat{x}_k$};

  % deviation arrow between Khat and K
  \draw[->, thick, black!80] (Khat) -- (K)
    node[midway, right=2pt, black!80] {$w(k{\to}t)$};

  % label cache rollout
  \node[red!75!black, anchor=south, font=\footnotesize]
    at (4.7, {3.0 - 0.20*4.7 - 0.018*4.7*4.7 + 0.55})
    {cached rollout using $\Delta\mathbf{h}_t,\Delta\mathbf{e}_t$};
\end{tikzpicture}
\caption{Cache deviation as edge weight. Starting from $x_t^*$ on the
ideal trajectory, a single full pass populates the cache with residuals
$\Delta\mathbf{h}_t,\Delta\mathbf{e}_t$. We then take $t{-}k$ Euler
updates that all reuse those residuals (red dashed), arriving at
$\hat{x}_k$. The squared distance to the ideal endpoint $x_k^*$ is the
cache-deviation edge weight $w(k\to t)$ used in our graph.}
\label{fig:cache_dev}
\end{figure}
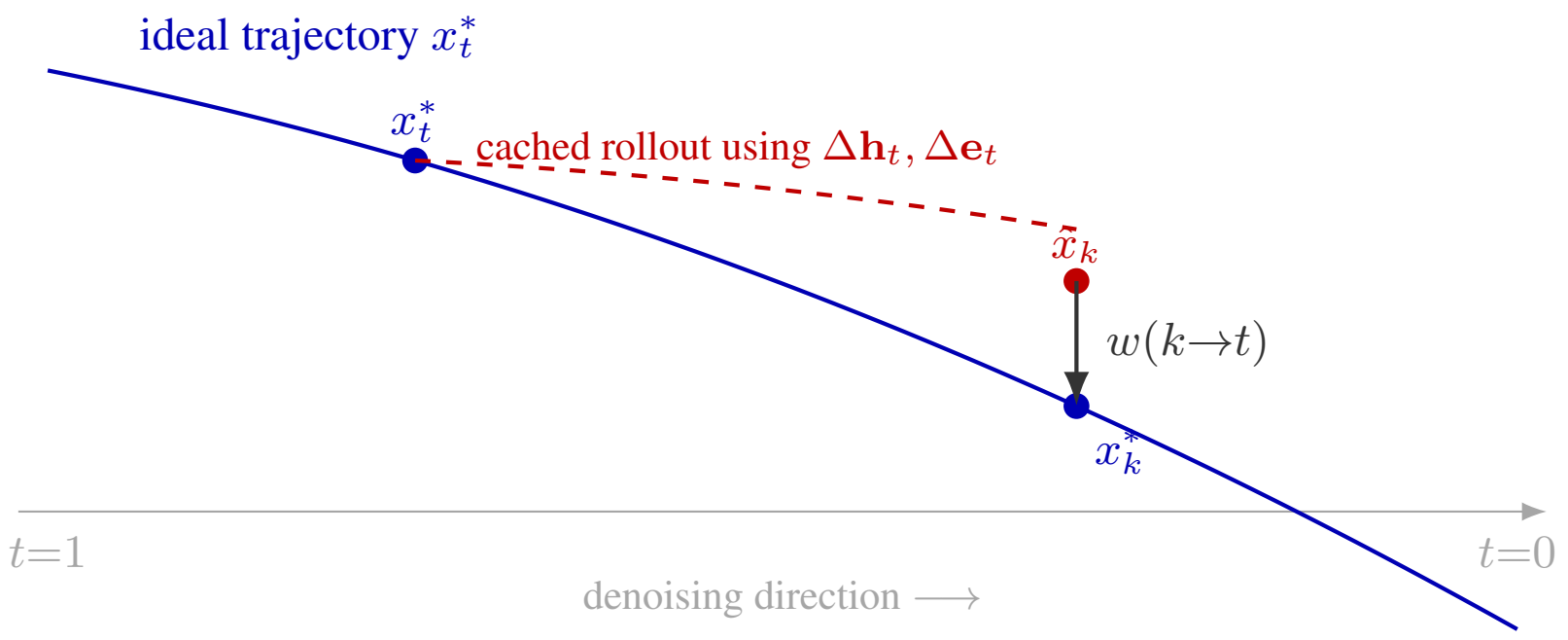

Let $\hat{x}_k$ denote the endpoint of this cached rollout. Since the ideal trajectory specifies the desired state $x_k^{*}$ at step $k$, we measure the deviation induced by caching from $t$ to $k$ as
\begin{equation}
    w(k\to t) \;=\;
    \bigl\|\,\mathrm{cache}(x_t^{*},\, k) - x_k^{*}\bigr\|^2
    \;=\;
    \bigl\|\hat{x}_k - x_k^{*}\bigr\|^2.
    \label{eq:cache_deviation}
\end{equation}
We refer to this quantity as the \emph{cache deviation}. It characterizes how well residuals computed at step $t$ generalize when reused over the interval $[k,t]$. It depends only on the cache contents at step $t$ and the corresponding ideal state at step $k$, and does not incorporate errors from earlier timesteps (Proposition~\ref{prop:additive}). Figure~\ref{fig:cache_dev} illustrates a single measurement of this deviation along the ideal denoising trajectory.

\noindent \textbf{What the edge weight measures.}
The two propositions below justify the edge weight and the additive path cost; proofs are given in Appendix~\ref{app:proofs}.

\begin{proposition}[The mean edge weight targets the marginal update]
\label{prop:marginal}
Let $f(x_t^{*}) = \mathrm{cache}(x_t^{*}, k)$ be the deterministic cached map. Then
\begin{equation}
\begin{aligned}
    \mathbb{E}\,\|f(x_t^{*}) - x_k^{*}\|^2
    &= \mathbb{E}\,\bigl\|f(x_t^{*}) - \mathbb{E}[x_k^{*} \mid x_t^{*}]\bigr\|^2
    + \mathbb{E}\,\mathrm{Var}(x_k^{*} \mid x_t^{*}), \\
    \mathbb{E}[x_k^{*} \mid x_t^{*}] &= x_t^{*} - (t-k)\,v^{\star},
\end{aligned}
\label{eq:prop_marginal}
\end{equation}
where $v^{\star} = \mathbb{E}[\epsilon - x_0^{*} \mid x_t^{*}]$ is the marginal velocity field that flow-matching models are trained to approximate~\citep{lipman2023flow,liu2023flow}, and the variance term does not depend on the schedule.
\end{proposition}

Hence, up to a schedule-independent constant, the mean edge weight (Eq.~\ref{eq:mean_edge}) measures the deviation of the cached rollout from the model's exact marginal update. A single linear interpolant is only an unbiased estimator of this target, and averaging over interpolants (which may intersect) is what makes it well defined. Once averaged, the objective coincides with the conditional flow matching objective~\citep{lipman2023flow,tong2024improving}; the difference is that flow matching uses it to train a network, whereas we keep the network fixed and use it to select a caching schedule.

\begin{proposition}[Per-interval deviation]
\label{prop:additive}
Assume linear flow, and let the cached velocity estimate be constant over $(k,t)$ and equal to $v_t = v_\theta(x_t^{*}, t, \mathbf{c})$, the model's velocity output at the full-evaluation step $t$. With $v^{\star} = \epsilon - x_0^{*}$,
\begin{equation}
    \hat{x}_k - x_k^{*} = (t-k)\,(v^{\star} - v_t),
    \qquad
    w(k \to t) = (t-k)^2\,\|v^{\star} - v_t\|^2.
    \label{eq:prop_additive}
\end{equation}
\end{proposition}

The deviation over an interval therefore depends only on the model's velocity error at $t$ and the interval length, and not on any earlier interval. Each cached interval incurs a single, self-contained error, which gives the optimal-substructure property that makes the additive path cost of Sec.~\ref{sec:graph} exact and its dynamic-programming solution optimal.

\noindent \textbf{Aggregating across runs and assembling the graph.}
\label{sec:graph_construction}
A key observation is that standard diffusion samplers typically rely on a fixed, hand-designed uniform 50-step schedule, which provides a stable discretization over $[0,1]$ across different prompts and noise realizations, making cache-induced deviations comparable across runs.
Given $R$ ideal trajectories, we obtain measurements $\{w^{(r)}(k\to t)\}_{r=1}^{R}$ for each pair of timesteps $(k,t)$ under different prompts and seeds. We define the mean edge weight as
\begin{equation}
    \bar{w}(k \to t) \;=\; \tfrac{1}{R}\sum_{r=1}^{R} w^{(r)}(k\to t),
    \label{eq:mean_edge}
\end{equation}
which induces a mean graph $\bar{\mathcal{G}}$ over timesteps.

This averaging yields an edge cost that reflects the expected denoising dynamics under the fixed schedule, rather than individual trajectory noise. The resulting graph $\bar{\mathcal{G}}$ is used as the optimization space in the next section.

\subsection{Graph-based optimal schedule search}
\label{sec:graph}

Given the mean graph $\bar{\mathcal{G}}$ constructed in Sec.~\ref{sec:graph_construction}, we now formulate the selection of an optimal caching schedule as a shortest path problem.

\noindent \textbf{Graph structure.}
The graph $\bar{\mathcal{G}}$ is a directed acyclic graph over candidate timesteps $0,\dots,M{-}1$. Each edge $k \to t$ ($k < t$) is assigned the mean cache deviation $\bar{w}(k \to t)$ defined in Eq.~\ref{eq:mean_edge}. To encode schedule selection as a path problem, we introduce a super-source $s$ and a super-sink $e$, following the standard construction in CoTj~\citep{chen2026cotj}. The resulting path from $s$ to $e$ corresponds to a valid selection of compute steps under budget $B$.

\noindent \textbf{Optimal schedule.}
We seek the minimum-weight path from the super-source $s$ to the super-sink $e$ that uses exactly $B$ real timestep nodes:
\begin{equation}
    \mathcal{P}^{*} =
        \underset{\mathcal{P}\,\in\,\mathcal{P}_{B+2}}{\arg\min}
        \;\sum_{(u,\,v)\,\in\,\mathcal{P}} \bar{w}(u \to v),
    \label{eq:shortest_path}
\end{equation}
which can be solved exactly via dynamic programming on the DAG:
\begin{equation}
    D[i][k] = \min_{j < i} \{ D[j][k{-}1] + \bar{w}(j \to i) \},
    \qquad D[i][1] = \bar{w}(s \to i).
    \label{eq:dp}
\end{equation}
The optimal cost is $\min_i \{ D[i][B] + \bar{w}(i \to e) \}$, and the corresponding schedule is recovered by backtracking. Here $\bar{w}(s \to i)$ and $\bar{w}(i \to e)$ denote the source and sink edge weights, respectively, which are defined based on the single-step reconstruction error at timestep $i$.
\subsection{Overview and Inference Pipeline}
The overall inference procedure is summarized in Algorithm~\ref{alg:main} in the Appendix~\ref{app:algorithm}. 
Given the optimal schedule $\mathcal{P}^{*} = (i_1,\dots,i_B)$, inference proceeds by executing full transformer evaluations only at selected timesteps $t_{i_j}$ while reusing cached residuals for all intermediate steps via Eq.~\ref{eq:cache_euler}. This replaces the standard uniform schedule with a budget-optimal, cache-aware trajectory without additional runtime overhead compared to existing caching-based samplers.

%%%%%%%%%%%%%%%%%%%%%%%%%%%%%%%%%%%%%%%%%%%%%%%%%%%%%%%%%%%%
\section{Experiments}
%%%%%%%%%%%%%%%%%%%%%%%%%%%%%%%%%%%%%%%%%%%%%%%%%%%%%%%%%%%%

\begin{table}[t]
\centering
\caption{Quantitative comparison of \textbf{image editing} on Qwen-Image-Edit, GEdit-Bench. (SC: Semantic Consistency, PQ: Perceptual Quality, OS: Overall Score.) Wall-clock speedup is the end-to-end latency ratio at batch size 1 and also includes the fixed cost of the VAE, text encoder, and scheduler; FLOPs speedup is the ratio of transformer FLOPs.}
\label{tab:image_editing}
\renewcommand{\arraystretch}{1.0}
\setlength{\tabcolsep}{4pt}
\small
\resizebox{\linewidth}{!}{%
\begin{tabular}{lcccccccccc}
\toprule
\multirow{2}{*}{\textbf{Method}}
  & \multicolumn{4}{c}{\textbf{Acceleration}}
  & \multicolumn{3}{c}{\textbf{GEdit-CN (Full)}}
  & \multicolumn{3}{c}{\textbf{GEdit-EN (Full)}} \\
\cmidrule(lr){2-5}\cmidrule(lr){6-8}\cmidrule(lr){9-11}
  & Latency(s)$\downarrow$ & Wall-clock Speed$\uparrow$
  & FLOPs(T)$\downarrow$ & FLOPs Speed$\uparrow$
  & SC$\uparrow$ & PQ$\uparrow$ & OS$\uparrow$
  & SC$\uparrow$ & PQ$\uparrow$ & OS$\uparrow$ \\
\midrule
\textbf{Vanilla: 50 steps}
  & 268.94 & 1.00$\times$ & 28190.88 & 1.00$\times$
  & 7.68 & 7.51 & 7.41 & 7.82 & 7.54 & 7.54 \\
50\% steps & 135.44 & 1.99$\times$ & 14095.44 & 2.00$\times$
  & 7.70 & 7.53 & 7.44 & 7.77 & 7.52 & 7.47 \\
20\% steps & 55.25 & 4.87$\times$ & 5638.18 & 5.00$\times$
  & 7.65 & 7.42 & 7.35 & 7.73 & 7.46 & 7.44 \\
\midrule
FORA ($\mathcal{N}{=}5$) & 59.71 & 4.51$\times$ & 5643.13 & 5.00$\times$
  & 7.60 & 7.31 & 7.25 & 7.62 & 7.34 & 7.28 \\
DuCa ($\mathcal{N}{=}6, R{=}90\%$) & 67.06 & 4.01$\times$ & 5897.67 & 4.78$\times$
  & 7.63 & 7.44 & 7.44 & 7.68 & 7.42 & 7.39 \\
TaylorSeer ($\mathcal{N}{=}6$) & 62.06 & 4.33$\times$ & 5643.13 & 5.00$\times$
  & 7.25 & 7.09 & 6.92 & 7.26 & 7.14 & 6.89 \\
HyCa ($\mathcal{N}{=}6$) & 59.44 & 4.52$\times$ & 5642.24 & 5.00$\times$
  & \textbf{7.76} & 7.49 & 7.50 & 7.77 & 7.47 & 7.45 \\
\rowcolor{myrow}
IdealCache ($\mathcal{B}{=}9$) & \textbf{56.62} & \textbf{4.75}$\times$
  & \textbf{5160.83} & \textbf{5.46}$\times$
  & 7.70 & \textbf{7.60} & 7.52
  & \textbf{7.86} & \textbf{7.59} & \textbf{7.62} \\
\midrule
FORA ($\mathcal{N}{=}7$) & 49.35 & 5.45$\times$ & 4515.74 & 6.24$\times$
  & 7.42 & 7.13 & 7.06 & 7.43 & 7.19 & 7.06 \\
DuCa ($\mathcal{N}{=}10, R{=}95\%$) & 56.55 & 4.76$\times$ & 5158.45 & 5.46$\times$
  & 7.50 & 5.75 & 6.39 & 7.52 & 5.77 & 6.41 \\
TaylorSeer ($\mathcal{N}{=}8$) & 50.98 & 5.28$\times$ & 4515.74 & 6.24$\times$
  & 6.61 & 6.65 & 6.31 & 6.67 & 6.63 & 6.31 \\
HyCa ($\mathcal{N}{=}8$) & 48.30 & 5.57$\times$ & 4514.48 & 6.24$\times$
  & \textbf{7.74} & 7.41 & 7.44 & 7.80 & 7.36 & 7.42 \\
\rowcolor{myrow}
IdealCache ($\mathcal{B}{=}7$) & \textbf{46.05} & \textbf{5.84}$\times$
  & \textbf{4037.43} & \textbf{6.98}$\times$
  & 7.70 & \textbf{7.58} & \textbf{7.51}
  & \textbf{7.87} & 7.55 & \textbf{7.59} \\
\bottomrule
\end{tabular}%
}
\end{table}
\begin{table}[t]
\centering
\caption{Results on the FlowEdit dataset. ``Consistency'' measures similarity to vanilla 50-step results, while ``Background Preservation'' evaluates background similarity to the input image using SSIM.}
\label{tab:flowedit}
\renewcommand{\arraystretch}{1.0}
\small
\begin{tabularx}{\linewidth}{l *{3}{>{\centering\arraybackslash}X}}
\toprule
& ImageReward$\uparrow$ & Consistency$\uparrow$ & Background Preservation$\uparrow$ \\
\midrule
50 steps (vanilla) & 1.054 & - & 0.665 \\
\rowcolor{myrow}
Ours ($\mathcal{B}{=}25$) & 1.049 & 0.960 & 0.687 \\
\rowcolor{myrow}
Ours ($\mathcal{B}{=}10$) & 0.991 & 0.890 & 0.679 \\
\bottomrule
\end{tabularx}
\end{table}
\subsection{Experimental Setup}

We evaluate IdealCache across text-to-image (Qwen-Image~\citep{wu2025qwenimage}), image editing (Qwen-Image-Edit), and text-to-video (HunyuanVideo~\citep{kong2024hunyuanvideo}) generation. Speedups are anchored to the standard $50$-step vanilla pipeline on a single NVIDIA H800 GPU. We compare against seven recent caching baselines. Unlike prior works parameterized by a fixed cache interval $\mathcal{N}$, IdealCache optimizes the placement of full transformer passes under a global step budget $\mathcal{B}$. Full implementation details, including datasets, metrics, and trajectory construction, are deferred to Appendix~\ref{app:exp_setup}.

\begin{figure}[t]
\centering
\includegraphics[width=0.9\linewidth]{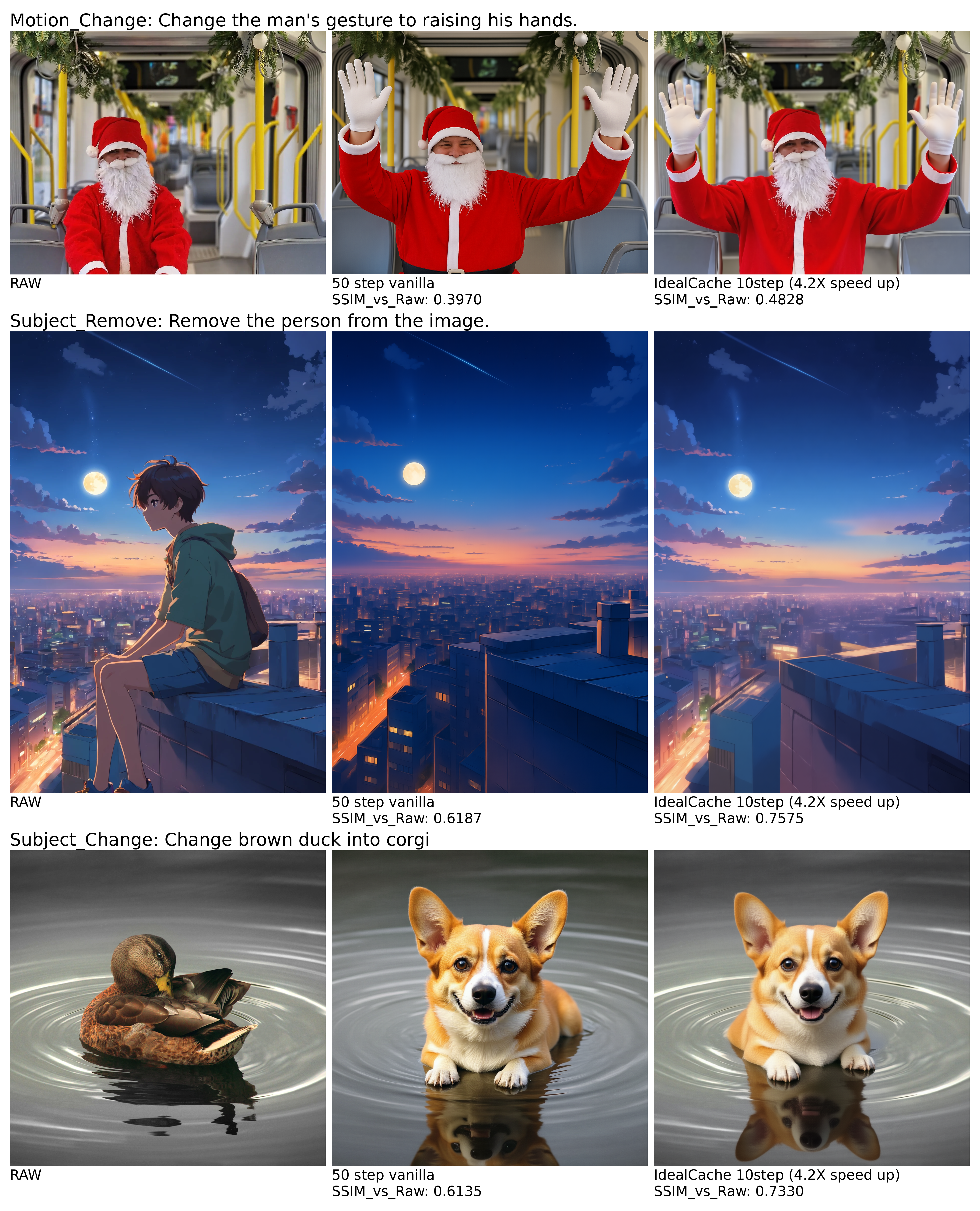}
\caption{Qualitative editing results on Qwen-Image-Edit. Our method achieves better background preservation in some cases while maintaining comparable editing quality. IdealCache with $\mathcal{B}{=}10$ ($\sim$$4.2\times$ speedup) achieves visual quality comparable to the vanilla 50-step baseline.}
\label{fig:flowedit_qual}
\end{figure}

\subsection{Results on image editing}
Following HyCA~\citep{zheng2025hyca}, we adopt the same evaluation setting for image editing. Table~\ref{tab:image_editing} reports the results on GEdit-Bench~\citep{liu2025step1x}. At $\mathcal{B}{=}9$, IdealCache achieves $4.75\times$ wall-clock speedup ($5.46\times$ FLOPs) with the best PQ and OS on both GEdit-CN and GEdit-EN, operating at lower FLOPs than HyCa ($\mathcal{N}{=}6$). Interestingly, on GEdit-EN, the optimised schedule's PQ and OS modestly exceed those of the unmodified $50$-step vanilla pipeline (e.g., OS $7.62$ vs. $7.54$). This highlights a subtle but important benefit of our formulation: anchoring the schedule on an ideal target rather than intermediate states allows the model to occasionally bypass approximation errors inherent in strictly tracking the reference trajectory.

At a tighter budget of $\mathcal{B}{=}7$ ($5.84\times$ speedup), IdealCache continues to outperform all caching baselines on PQ and OS. Table~\ref{tab:flowedit} further explores this behaviour on the FlowEdit~\citep{kulikov2024flowedit} dataset, taking advantage of the unique nature of image editing tasks. In editing, preserving the unedited background is a primary objective. By directly setting the source image as the ideal target $x_0^{\star}$ in our trajectory construction, our graph inherently penalises background drift during the shortest-path search. As a result, the recovered schedule naturally traces a functionally cleaner path, achieving strictly superior background preservation compared to the full $50$-step vanilla pipeline ($0.687$ and $0.679$ vs. $0.665$) while maintaining competitive ImageReward at massive speedups. Additional qualitative results in Figure~\ref{fig:flowedit_qual} visually confirm this improved background consistency.
% Table~\ref{tab:image_editing} presents results on GEdit-Bench. At
% $\mathcal{B}{=}9$, IdealCache achieves $4.80\times$ wall-clock speedup
% ($5.46\times$ FLOPs) with the best PQ and OS on both GEdit-CN
% (PQ~$7.60$, OS~$7.52$) and GEdit-EN (PQ~$7.59$, OS~$7.62$), at lower
% FLOPs than HyCa ($\mathcal{N}{=}6$). Notably on GEdit-EN, the optimised
% schedule's PQ and OS modestly exceed those of the unmodified $50$-step
% vanilla pipeline (PQ~$7.54$, OS~$7.54$), indicating that anchoring on
% the ideal trajectory rather than on the vanilla pipeline's intermediate
% states is not structurally bounded by the pipeline's output. At a
% tighter budget, $\mathcal{B}{=}7$ delivers $5.84\times$ wall-clock
% speedup ($6.98\times$ FLOPs) with PQ~$7.58$ and OS~$7.51$ on GEdit-CN
% (PQ~$7.55$, OS~$7.59$ on GEdit-EN), still outperforming all caching
% baselines on PQ and OS.
% Table~\ref{tab:flowedit} further confirms on the
% FlowEdit~\citep{kulikov2024flowedit} benchmark that our cache-based
% schedule matches the ImageReward of the vanilla pipeline at the same
% budget while preserving more of the intended edit.
% Additional qualitative results are shown in Figure~\ref{fig:flowedit_qual}.

\begin{table}[h]
\centering
\caption{Quantitative comparison of \textbf{text-to-image generation} on Qwen-Image, evaluated on the DrawBench benchmark. Wall-clock speedup is the end-to-end latency ratio at batch size 1 and also includes the fixed cost of the VAE, text encoder, and scheduler; FLOPs speedup is the ratio of transformer FLOPs.}
\label{tab:t2i}
\renewcommand{\arraystretch}{1.0}
\setlength{\tabcolsep}{4pt}
\small
\resizebox{\linewidth}{!}{%
\begin{tabular}{lccccccc}
\toprule
\multirow{2}{*}{\textbf{Method}}
  & \multicolumn{4}{c}{\textbf{Acceleration}}
  & \textbf{Quality}
  & \multicolumn{2}{c}{\textbf{Perceptual}} \\
\cmidrule(lr){2-5}\cmidrule(lr){6-6}\cmidrule(lr){7-8}
  & Latency(s)$\downarrow$ & Wall-clock Speed$\uparrow$
  & FLOPs(T)$\downarrow$ & FLOPs Speed$\uparrow$
  & ImageReward$\uparrow$
  & SSIM$\uparrow$ & LPIPS$\downarrow$ \\
\midrule
\textbf{Vanilla: 50 steps}
  & 46.54 & 1.00$\times$ & 10925.18 & 1.00$\times$
  & 1.2092 (+0.00\%) & --- & --- \\
50\% steps & 23.44 & 1.99$\times$ & 5462.59 & 2.00$\times$
  & 1.1612 ($-$3.97\%) & 0.798 & 0.249 \\
20\% steps & 9.51 & 4.89$\times$ & 2184.56 & 5.00$\times$
  & 0.8996 ($-$25.60\%) & 0.627 & 0.504 \\
\midrule
TaylorSeer ($\mathcal{N}{=}3$) & 22.92 & 2.03$\times$ & 3929.69 & 2.78$\times$
  & 1.0298 ($-$14.83\%) & 0.504 & 0.628 \\
HyCa ($\mathcal{N}{=}3$) & 21.95 & 2.12$\times$ & 3929.69 & 2.78$\times$
  & \textbf{1.1914} ($-$1.47\%) & 0.763 & 0.247 \\
\rowcolor{myrow}
IdealCache ($\mathcal{B}{=}17$) & \textbf{16.07} & \textbf{2.90}$\times$
  & \textbf{3727.28} & \textbf{2.93}$\times$
  & 1.1767 ($-$2.69\%) & 0.830 & 0.213 \\
\midrule
FORA ($\mathcal{N}{=}5$) & 13.49 & 3.45$\times$ & 2186.17 & 5.00$\times$
  & 0.7485 ($-$38.10\%) & 0.553 & 0.556 \\
ToCa ($\mathcal{N}{=}8$) & 37.66 & 1.24$\times$ & 2529.19 & 4.32$\times$
  & 0.9322 ($-$22.91\%) & 0.643 & 0.417 \\
DuCa ($\mathcal{N}{=}9$) & 15.43 & 3.02$\times$ & 2501.09 & 4.37$\times$
  & 0.7916 ($-$34.54\%) & 0.582 & 0.531 \\
TaylorSeer ($\mathcal{N}{=}6$) & 15.29 & 3.04$\times$ & 2186.17 & 5.00$\times$
  & 0.9139 ($-$24.42\%) & 0.504 & 0.628 \\
HyCa ($\mathcal{N}{=}6$) & 13.41 & 3.47$\times$ & 2185.32 & 5.00$\times$
  & 1.1506 ($-$4.85\%) & 0.709 & 0.320 \\
\rowcolor{myrow}
IdealCache ($\mathcal{B}{=}9$) & \textbf{8.98} & \textbf{5.18}$\times$
  & \textbf{1982.33} & \textbf{5.51}$\times$
  & \textbf{1.1814} ($-$2.30\%) & 0.755 & 0.327 \\
\midrule
FORA ($\mathcal{N}{=}6$) & 11.11 & 4.19$\times$ & 1964.93 & 5.56$\times$
  & 0.4608 ($-$61.91\%) & 0.546 & 0.597 \\
ToCa ($\mathcal{N}{=}12$) & 32.75 & 1.42$\times$ & 2035.04 & 5.37$\times$
  & 0.5391 ($-$55.42\%) & 0.589 & 0.519 \\
DuCa ($\mathcal{N}{=}12$) & 13.55 & 3.43$\times$ & 1836.61 & 5.95$\times$
  & 0.5036 ($-$58.37\%) & 0.576 & 0.593 \\
TaylorSeer ($\mathcal{N}{=}7$) & 10.77 & 4.32$\times$ & 1964.93 & 5.56$\times$
  & 0.8783 ($-$27.37\%) & 0.481 & 0.652 \\
HyCa ($\mathcal{N}{=}8$) & 8.65 & 5.38$\times$ & 1747.70 & 6.25$\times$
  & 1.0419 ($-$13.83\%) & 0.600 & 0.433 \\
\rowcolor{myrow}
IdealCache ($\mathcal{B}{=}7$) & \textbf{7.33} & \textbf{6.35}$\times$
  & \textbf{1546.10} & \textbf{7.07}$\times$
  & \textbf{1.1068} ($-$8.47\%) & 0.666 & 0.457 \\
\bottomrule
\end{tabular}%
}
\end{table}

\subsection{Results on text-to-image generation}

Table~\ref{tab:t2i} reports results on Qwen-Image using prompts from
DrawBench~\citep{saharia2022photorealistic}. Our method consistently
outperforms all caching baselines across every speed
group. At $\mathcal{B}{=}17$, IdealCache achieves $2.90\times$
wall-clock speedup ($2.93\times$ FLOPs) with SSIM $0.830$ and
ImageReward $1.1767$ ($-2.69\%$); at the same quality level, HyCa
($\mathcal{N}{=}3$) delivers only $2.12\times$ at SSIM $0.763$. At
$\mathcal{B}{=}9$, IdealCache reaches $5.18\times$ wall-clock speedup
($5.51\times$ FLOPs) with the best ImageReward $1.1814$ ($-2.30\%$) and
SSIM $0.755$ at this compression. At the highest compression,
$\mathcal{B}{=}7$ delivers $6.35\times$ wall-clock speedup
($7.07\times$ FLOPs) with ImageReward $1.1068$ ($-8.47\%$) and SSIM
$0.666$, substantially outperforming HyCa $\mathcal{N}{=}8$ at a
similar budget (ImageReward $1.0419$, $-13.83\%$). The vanilla
pipeline at $50$ steps remains the upper bound on aggregate
ImageReward; the quality--budget frontier traced by IdealCache strictly
Pareto-dominates all caching baselines at every compression level we
evaluate. We further note that the candidate-grid size~$M$ is a property of the
search, not of the sampler, and can be increased to improve quality
at no inference-cost penalty; see Appendix~\ref{app:grid_density}.

\subsection{Results on text-to-video generation}

Table~\ref{tab:t2v} reports results on HunyuanVideo evaluated at Vbench\citep{huang2024vbench} dataset at 61 frames, 320x512 resolution. At
$\mathcal{B}{=}14$, IdealCache achieves $3.24\times$ wall-clock speedup
with the best LPIPS ($0.159$) and SSIM ($0.817$) at this latency tier,
outperforming all caching baselines at comparable budgets. At
$\mathcal{B}{=}10$, the schedule reaches $4.32\times$ wall-clock speedup
(latency $18.74$s) with LPIPS $0.206$ and SSIM $0.760$, remaining
competitive with the strongest baselines. More qualitative results are provided in Appendix~\ref{app:gallery}.

\begin{table}[t]
\centering
\caption{Quantitative comparison of \textbf{text-to-video generation} on HunyuanVideo, evaluated on VBench. Wall-clock and FLOPs speedups are defined as in Table~\ref{tab:image_editing}; FLOPs are reported for the vanilla pipeline and IdealCache.}
\label{tab:t2v}
\renewcommand{\arraystretch}{1.0}
\setlength{\tabcolsep}{4pt}
\small
\resizebox{\linewidth}{!}{%
\begin{tabular}{lcccccc}
\toprule
\multirow{2}{*}{\textbf{Method}}
  & \multicolumn{4}{c}{\textbf{Acceleration}}
  & \multicolumn{2}{c}{\textbf{Visual Quality}} \\
\cmidrule(lr){2-5}\cmidrule(lr){6-7}
  & Latency(s)$\downarrow$ & Wall-clock Speed$\uparrow$
  & FLOPs(T)$\downarrow$ & FLOPs Speed$\uparrow$
  & LPIPS$\downarrow$ & SSIM$\uparrow$ \\
\midrule
\textbf{Vanilla: 50 steps} & 80.94 & 1.00$\times$ & 11561.3 & 1.00$\times$ & --- & --- \\
30\% steps & 30.21 & 2.68$\times$ & --- & --- & 0.381 & 0.659 \\
\midrule
ToCa ($\mathcal{N}{=}5$) & 27.64 & 2.93$\times$ & --- & --- & 0.454 & 0.590 \\
DuCa ($\mathcal{N}{=}5$) & 26.23 & 3.09$\times$ & --- & --- & 0.454 & 0.595 \\
DiCache ($\delta{=}0.8$) & 26.03 & 3.11$\times$ & --- & --- & 0.382 & 0.701 \\
TaylorSeer ($\mathcal{N}{=}5, O{=}1$) & 26.71 & 3.03$\times$ & --- & --- & 0.428 & 0.603 \\
TeaCache ($l{=}0.33$) & 26.03 & 3.11$\times$ & --- & --- & 0.363 & 0.651 \\
\rowcolor{myrow}
Ours ($\mathcal{B}{=}14$) & \textbf{24.97} & \textbf{3.24}$\times$
  & 3502.1 & 3.30$\times$
  & \textbf{0.159} & \textbf{0.817} \\
\midrule
DiCache ($\delta{=}3.0$) & 24.31 & 3.33$\times$ & --- & --- & 0.583 & 0.490 \\
TeaCache ($l{=}0.39$) & 24.35 & 3.32$\times$ & --- & --- & 0.396 & 0.631 \\
TaylorSeer ($\mathcal{N}{=}7, O{=}1$) & 24.07 & 3.36$\times$ & --- & --- & 0.480 & 0.595 \\
\rowcolor{myrow}
Ours ($\mathcal{B}{=}10$) & \textbf{18.74} & \textbf{4.32}$\times$
  & 2606.7 & 4.44$\times$
  & \textbf{0.206} & \textbf{0.760} \\
\bottomrule
\end{tabular}%
}
\end{table}

% \begin{table}[t]
% \centering
% \caption{Quantitative comparison of \textbf{text-to-video generation} on HunyuanVideo.}
% \label{tab:t2v}
% \renewcommand{\arraystretch}{1.0}
% \setlength{\tabcolsep}{4pt}
% \small
% \begin{tabular}{lcccc}
% \toprule
% \multirow{2}{*}{\textbf{Method} (\textit{HunyuanVideo})}
%   & \multicolumn{2}{c}{\textbf{Acceleration}}
%   & \multicolumn{2}{c}{\textbf{Visual Quality}} \\
% \cmidrule(lr){2-3}\cmidrule(lr){4-5}
%   & Latency(s)$\downarrow$ & Speed$\uparrow$
%   & LPIPS$\downarrow$ & SSIM$\uparrow$ \\
% \midrule
% \textbf{Vanilla: 50 steps} & 80.94 & 1.00$\times$ & --- & --- \\
% 30\% steps & 30.21 & 2.68$\times$ & 0.381 & 0.659 \\
% \midrule
% ToCa ($\mathcal{N}{=}5$) & 27.64 & 2.93$\times$ & 0.454 & 0.590 \\
% DuCa ($\mathcal{N}{=}5$) & 26.23 & 3.09$\times$ & 0.454 & 0.595 \\
% DiCache ($\delta{=}0.8$) & 25.80 & 3.11$\times$ & 0.382 & 0.701 \\
% TaylorSeer ($\mathcal{N}{=}5, O{=}1$) & 26.71 & 3.03$\times$ & 0.428 & 0.603 \\
% TeaCache ($l{=}0.33$) & 26.03 & 3.11$\times$ & 0.363 & 0.651 \\
% \rowcolor{myrow}
% Ours ($\mathcal{B}{=}14$) & \textbf{24.97} & \textbf{3.24}$\times$
%   & \textbf{0.159} & \textbf{0.817} \\
% \midrule
% DiCache ($\delta{=}3.0$) & 24.31 & 3.33$\times$ & 0.583 & 0.490 \\
% TeaCache ($l{=}0.39$) & 24.35 & 3.32$\times$ & 0.396 & 0.631 \\
% TaylorSeer ($\mathcal{N}{=}7, O{=}1$) & 24.07 & 3.36$\times$ & 0.480 & 0.595 \\
% \rowcolor{myrow}
% Ours ($\mathcal{B}{=}10$) & \textbf{18.74} & \textbf{4.32}$\times$
%   & \textbf{0.206} & \textbf{0.760} \\
% \bottomrule
% \end{tabular}
% \end{table}

\section{Minimal offline cost via out-of-distribution transferability}
\label{sec:ablation_ood}

Our main results rely on mean graphs aggregated over an ensemble of reference runs ($R{=}1000$). To evaluate the practical efficiency of IdealCache, we investigate the extreme \emph{one-shot} setting ($R{=}1$), where a schedule is solved from a single random prompt/image and applied to the entire dataset.

As shown in Table~\ref{tab:ood_transfer}, the one-shot schedules achieve performance nearly identical to the heavy ensemble across both tasks. This indicates that the selected schedule is largely insensitive to the particular prompt used to build the graph, so that a single reference run suffices in practice. Consequently, the offline cost of IdealCache can be reduced to an extremely low level: constructing a $50$-node graph from a single prompt takes approximately $80$ seconds, yet the resulting schedule generalizes effectively across the entire data distribution.
\begin{table}[t]
\centering
\caption{Out-of-distribution transferability on Qwen-Image and Qwen-Image-Edit ($\mathcal{B}{=}9$). We compare schedules derived from a single random prompt ($R{=}1$) against the heavy ensemble mean graph. The performance gap is negligible, demonstrating that extremely low-cost offline planning is highly viable.}
\label{tab:ood_transfer}
\renewcommand{\arraystretch}{1.0}
\small
\begin{tabularx}{\linewidth}{l *{6}{>{\centering\arraybackslash}X}}
\toprule
\multirow{2}{*}{\textbf{Graph Construction}} & \multicolumn{3}{c}{\textbf{Text-to-Image} (DrawBench)} & \multicolumn{3}{c}{\textbf{Image Editing} (GEdit-EN)} \\
\cmidrule(lr){2-4} \cmidrule(lr){5-7}
& SSIM$\uparrow$ & LPIPS$\downarrow$ & ImageReward$\uparrow$ & SC$\uparrow$ & PQ$\uparrow$ & OS$\uparrow$ \\
\midrule
Mean Graph ($R{=}1000$) & 0.755 & 0.327 & 1.181 & 7.86 & 7.59 & 7.62 \\
\rowcolor{myrow}
One-Shot ($R{=}1$) & 0.752 & 0.334 & 1.168 & 7.81 & 7.58 & 7.58 \\
\bottomrule
\end{tabularx}
\end{table}

\FloatBarrier
\section{Conclusion}
%%%%%%%%%%%%%%%%%%%%%%%%%%%%%%%%%%%%%%%%%%%%%%%%%%%%%%%%%%%%

We presented IdealCache, a training-free framework that abandons step-wise
imitation of the reference denoising trajectory and instead anchors
cache-based scheduling on a per-input ensemble of ideal trajectories
constructed from candidate target results, aggregated into a single
graph. By solving a constrained shortest path problem on this mean
graph --- whose edge costs measure the average deviation of cached
rollouts from the ensemble --- our method discovers non-uniform
schedules under any step budget. Because the optimisation target is the
ideal-trajectory geometry rather than the vanilla pipeline's
intermediate states, the schedule is freed from the structural bound of
the pipeline's output: in image editing the optimised schedules can
match and on certain task-specific metrics modestly exceed the
$50$-step vanilla pipeline, while across all three modalities they
Pareto-dominate prior caching baselines on the quality--budget
frontier. Experiments across text-to-image, image editing, and
text-to-video confirm broad applicability and strong
quality--efficiency trade-offs across diverse tasks and models.

%%%%%%%%%%%%%%%%%%%%%%%%%%%%%%%%%%%%%%%%%%%%%%%%%%%%%%%%%%%%
% \section*{References}
%%%%%%%%%%%%%%%%%%%%%%%%%%%%%%%%%%%%%%%%%%%%%%%%%%%%%%%%%%%%

\begin{ack}
This work received no third-party funding. The authors declare no competing interests.
\end{ack}

\bibliographystyle{plainnat}

\begin{thebibliography}{99}

\bibitem[Bu et al.(2025)]{bu2025dicache}
Bu, J., Ling, P., Zhou, Y., Wang, Y., Zang, Y., Wu, T., Lin, D., and
Wang, J.
\newblock DiCache: Let diffusion model determine its own cache.
\newblock \emph{arXiv:2508.17356}, 2025.

\bibitem[Chen et al.(2024a)]{chen2024trajectory}
Chen, D., Zhou, Z., Wang, C., Shen, C., and Lyu, S.
\newblock On the trajectory regularity of ODE-based diffusion sampling.
\newblock \emph{ICML}, 2024.

\bibitem[Chen et al.(2024b)]{chen2024deltadit}
Chen, P., Shen, M., Ye, P., Cao, J., Tu, C., Bouganis, C.-S., Zhao, Y.,
and Chen, T.
\newblock $\Delta$-DiT: A training-free acceleration method tailored for
diffusion transformers.
\newblock \emph{arXiv:2406.01125}, 2024.

\bibitem[Chen et al.(2025)]{chen2025shortdf}
Chen, P., Zhang, X., Liu, Z., Hu, H., Liu, X., Wang, K., Wang, M., Qian, Y.,
and Lian, S.
\newblock Optimizing for the shortest path in denoising diffusion model.
\newblock \emph{CVPR}, 2025.

\bibitem[Huang et al.(2024)]{huang2024vbench}
Huang, Z., He, Y., Yu, J., Zhang, F., Si, C., Jiang, Y., et al.
\newblock VBench: Comprehensive benchmark suite for video generative models.
\newblock \emph{CVPR}, 2024.

\bibitem[Chen et al.(2026)]{chen2026cotj}
Chen, P., Liu, X., Zhang, X., Shen, F., Gong, X., Liu, Z., Chen, Z.,
Hu, H., Wang, K., and Lian, S.
\newblock Chain-of-Trajectories: Unlocking the intrinsic generative
optimality of diffusion models via graph-theoretic planning.
\newblock \emph{arXiv:2603.14704}, 2026.

\bibitem[Kahatapitiya et al.(2024)]{kahatapitiya2024adacache}
Kahatapitiya, K., Liu, H., He, S., Liu, D., Jia, M., Zhang, C., Ryoo,
M.~S., and Xie, T.
\newblock Adaptive caching for faster video generation with diffusion
transformers.
\newblock \emph{arXiv:2411.02397}, 2024.

\bibitem[Karras et al.(2022)]{karras2022elucidating}
Karras, T., Aittala, M., Aila, T., and Laine, S.
\newblock Elucidating the design space of diffusion-based generative models.
\newblock \emph{NeurIPS}, 2022.

\bibitem[Kirstain et al.(2023)]{kirstain2023pickapic}
Kirstain, Y., Polyak, A., Singer, U., Matiana, S., Penna, J., and Levy, O.
\newblock Pick-a-Pic: An open dataset of user preferences for
text-to-image generation.
\newblock \emph{NeurIPS}, 2023.

\bibitem[Kong et al.(2024)]{kong2024hunyuanvideo}
Kong, W., Tian, Q., Zhang, Z., Min, R., Dai, Z., Zhou, J., et al.
\newblock HunyuanVideo: A systematic framework for large video generative models.
\newblock \emph{arXiv:2412.03603}, 2024.

\bibitem[Kulikov et al.(2025)]{kulikov2024flowedit}
Kulikov, V., Kleiner, M., Huberman-Spiegelglas, I., and Michaeli, T.
\newblock FlowEdit: Inversion-free text-based editing using pre-trained flow
models.
\newblock \emph{ICCV}, 2025.

\bibitem[Liu et al.(2024a)]{liu2024teacache}
Liu, F., Zhang, S., Wang, X., Wei, Y., Qiu, H., Zhao, Y., Zhang, Y., Ye, Q.,
and Wan, F.
\newblock Timestep embedding tells: It's time to cache for video diffusion model.
\newblock \emph{arXiv:2411.19108}, 2024.

\bibitem[Lipman et al.(2023)]{lipman2023flow}
Lipman, Y., Chen, R.~T.~Q., Ben-Hamu, H., Nickel, M., and Le, M.
\newblock Flow matching for generative modeling.
\newblock \emph{ICLR}, 2023.

\bibitem[Liu et al.(2025c)]{liu2025speca}
Liu, J., Zou, C., Lyu, Y., Ren, F., Wang, S., Li, K., and Zhang, L.
\newblock SpeCa: Accelerating diffusion transformers with speculative
feature caching.
\newblock \emph{ACM Multimedia}, 2025.

\bibitem[Liu et al.(2023a)]{liu2023flow}
Liu, X., Gong, C., and Liu, Q.
\newblock Flow straight and fast: Learning to generate and transfer data with
rectified flow.
\newblock \emph{ICLR}, 2023.

\bibitem[Liu et al.(2024b)]{liu2024instaflow}
Liu, X., Zhang, X., Ma, J., Peng, J., and Liu, Q.
\newblock InstaFlow: One step is enough for high-quality diffusion-based
text-to-image generation.
\newblock \emph{ICLR}, 2024.

\bibitem[Liu et al.(2025a)]{liu2025taylorseer}
Liu, J., Zou, C., Lyu, Y., Chen, J., and Zhang, L.
\newblock From reusing to forecasting: Accelerating diffusion models with
TaylorSeers.
\newblock \emph{ICCV}, 2025.

\bibitem[Liu et al.(2025b)]{liu2025step1x}
Liu, S., Han, Y., Xing, P., Yin, F., Wang, R., Cheng, W., et al.
\newblock Step1X-Edit: A practical framework for general image editing.
\newblock \emph{arXiv:2504.17761}, 2025.

\bibitem[Lu et al.(2022a)]{lu2022dpm}
Lu, C., Zhou, Y., Bao, F., Chen, J., Li, C., and Zhu, J.
\newblock DPM-Solver: A fast ODE solver for diffusion probabilistic model
sampling.
\newblock \emph{NeurIPS}, 2022.

\bibitem[Lu et al.(2022b)]{lu2022dpmpp}
Lu, C., Zhou, Y., Bao, F., Chen, J., Li, C., and Zhu, J.
\newblock DPM-Solver++: Fast solver for guided sampling of diffusion
probabilistic models.
\newblock \emph{arXiv:2211.01095}, 2022.

\bibitem[Luo et al.(2023)]{luo2023lcm}
Luo, S., Tan, Y., Huang, L., Li, J., and Zhao, H.
\newblock Latent consistency models: Synthesizing high-resolution images with
few-step inference.
\newblock \emph{arXiv:2310.04378}, 2023.

\bibitem[Lv et al.(2025)]{lv2025fastercache}
Lv, Z., Si, C., Song, J., Yang, Z., Qiao, Y., Liu, Z., and Wong, K.-Y.~K.
\newblock FasterCache: Training-free video diffusion model acceleration with
high quality.
\newblock \emph{ICLR}, 2025.

\bibitem[Ma et al.(2024a)]{ma2024deepcache}
Ma, X., Fang, G., and Wang, X.
\newblock DeepCache: Accelerating diffusion models for free.
\newblock \emph{CVPR}, 2024.

\bibitem[Ma et al.(2024b)]{ma2024ltc}
Ma, X., Fang, G., Mi, M.~B., and Wang, X.
\newblock Learning-to-cache: Accelerating diffusion transformer via layer
caching.
\newblock \emph{NeurIPS}, 2024.

\bibitem[Peebles \& Xie(2023)]{peebles2023dit}
Peebles, W. and Xie, S.
\newblock Scalable diffusion models with transformers.
\newblock \emph{ICCV}, 2023.

\bibitem[Sabour et al.(2024)]{sabour2024align}
Sabour, A., Fidler, S., and Kreis, K.
\newblock Align your steps: Optimizing sampling schedules in diffusion models.
\newblock \emph{ICML}, 2024.

\bibitem[Peng et al.(2026)]{peng2026ertacache}
Peng, X., Yan, C., Liu, H., Ma, R., Chen, F., Wang, X., Wu, Z., Liu, S.,
and Lin, M.
\newblock ERTACache: Error rectification and timesteps adjustment for
efficient diffusion.
\newblock \emph{ICLR}, 2026.

\bibitem[Qiu et al.(2025)]{qiu2025goc}
Qiu, J., Liu, L., Wang, S., Lu, J., Chen, K., and Hao, Y.
\newblock Accelerating diffusion transformer via gradient-optimized cache.
\newblock \emph{ICCV}, 2025.

\bibitem[Saharia et al.(2022)]{saharia2022photorealistic}
Saharia, C., Chan, W., Saxena, S., Li, L., Whang, J., Denton, E., et al.
\newblock Photorealistic text-to-image diffusion models with deep
language understanding.
\newblock \emph{NeurIPS}, 2022.

\bibitem[Salimans \& Ho(2022)]{salimans2022distillation}
Salimans, T. and Ho, J.
\newblock Progressive distillation for fast sampling of diffusion models.
\newblock \emph{ICLR}, 2022.

\bibitem[Selvaraju et al.(2024)]{selvaraju2024fora}
Selvaraju, P., Ding, T., Chen, T., Zharkov, I., and Liang, L.
\newblock FORA: Fast-forward caching in diffusion transformer acceleration.
\newblock \emph{arXiv:2407.01425}, 2024.

\bibitem[Song et al.(2021)]{song2021ddim}
Song, J., Meng, C., and Ermon, S.
\newblock Denoising diffusion implicit models.
\newblock \emph{ICLR}, 2021.

\bibitem[Song et al.(2023)]{song2023consistency}
Song, Y., Dhariwal, P., Chen, M., and Sutskever, I.
\newblock Consistency models.
\newblock \emph{ICML}, 2023.

\bibitem[Sun et al.(2026)]{sun2026jit}
Sun, W., Li, J., and Liu, Z.
\newblock Just-in-Time: Training-free spatial acceleration for diffusion
transformers.
\newblock \emph{CVPR}, 2026.

\bibitem[Tong et al.(2024)]{tong2024improving}
Tong, A., Fatras, K., Malkin, N., Huguet, G., Zhang, Y., Rector-Brooks, J.,
Wolf, G., and Bengio, Y.
\newblock Improving and generalizing flow-based generative models with
minibatch optimal transport.
\newblock \emph{TMLR}, 2024.

\bibitem[Wang et al.(2004)]{wang2004ssim}
Wang, Z., Bovik, A.~C., Sheikh, H.~R., and Simoncelli, E.~P.
\newblock Image quality assessment: from error visibility to structural
similarity.
\newblock \emph{IEEE Transactions on Image Processing}, 2004.

\bibitem[Wu et al.(2025)]{wu2025qwenimage}
Wu, C., Li, J., Zhou, J., Lin, J., Gao, K., Yan, K., et al.
\newblock Qwen-Image technical report.
\newblock \emph{arXiv:2508.02324}, 2025.

\bibitem[Xu et al.(2023)]{xu2023imagereward}
Xu, J., Liu, X., Wu, Y., Tong, Y., Li, Q., Ding, M., Tang, J., and Dong, Y.
\newblock ImageReward: Learning and evaluating human preferences for
text-to-image generation.
\newblock \emph{NeurIPS}, 2023.

\bibitem[Zhang \& Chen(2023)]{zhang2023deis}
Zhang, Q. and Chen, Y.
\newblock Fast sampling of diffusion models with exponential integrator.
\newblock \emph{ICLR}, 2023.

\bibitem[Zhang et al.(2018)]{zhang2018lpips}
Zhang, R., Isola, P., Efros, A.~A., Shechtman, E., and Wang, O.
\newblock The unreasonable effectiveness of deep features as a
perceptual metric.
\newblock \emph{CVPR}, 2018.

\bibitem[Zhao et al.(2023)]{zhao2023unipc}
Zhao, W., Bai, L., Rao, Y., Zhou, J., and Lu, J.
\newblock UniPC: A unified predictor-corrector framework for fast sampling of
diffusion models.
\newblock \emph{NeurIPS}, 2023.

\bibitem[Zhao et al.(2024)]{zhao2024pab}
Zhao, X., Jin, X., Wang, K., and You, Y.
\newblock Real-time video generation with pyramid attention broadcast.
\newblock \emph{arXiv:2408.12588}, 2024.

\bibitem[Zheng et al.(2025)]{zheng2025hyca}
Zheng, S., Chen, G., Zhou, Q., Lin, Y., He, L., Zou, C., Cai, P., Liu, J.,
and Zhang, L.
\newblock Let features decide their own solvers: Hybrid feature caching for
diffusion transformers.
\newblock \emph{arXiv:2510.04188}, 2025.

\bibitem[Zou et al.(2024a)]{zou2024toca}
Zou, C., Liu, X., Liu, T., Huang, S., and Zhang, L.
\newblock Accelerating diffusion transformers with token-wise feature caching.
\newblock \emph{ICLR}, 2025.

\bibitem[Zou et al.(2024b)]{zou2024duca}
Zou, C., Zhang, E., Guo, R., Xu, H., He, C., Hu, X., and Zhang, L.
\newblock Rethinking token-wise feature caching: Accelerating diffusion
transformers with dual feature caching.
\newblock \emph{arXiv:2412.18911}, 2024.

\end{thebibliography}

%%%%%%%%%%%%%%%%%%%%%%%%%%%%%%%%%%%%%%%%%%%%%%%%%%%%%%%%%%%%
\clearpage
\appendix
%%%%%%%%%%%%%%%%%%%%%%%%%%%%%%%%%%%%%%%%%%%%%%%%%%%%%%%%%%%%

\section{Algorithm summary and cost analysis}
\label{app:algorithm}

This appendix provides a full pseudocode summary of IdealCache
(Algorithm~\ref{alg:main}) and a wall-clock cost analysis.

\begin{algorithm}[h]
\caption{\textbf{IdealCache}: Ideal-Trajectory-Anchored Optimal Schedule Search}
\label{alg:main}
\begin{algorithmic}[1]
\small
\Require Conditioning $\mathbf{c}$, candidate-grid size $M$, step budget $B$,
         number of reference runs $R$
\Ensure Schedule $\mathcal{P}^{*}$, output $\hat{x}_0$
\State \textbf{// Phase 1: Target-Result Construction \& Ideal Trajectories}
\State Initialise $\bar{w}(\cdot) \leftarrow 0$
\For{$r = 1$ \textbf{to} $R$}
    \State $x_0^{*(r)} \leftarrow \texttt{FullDenoise}(\mathbf{c},\, M,\,
           \epsilon^{(r)})$
    \For{$i = 1$ \textbf{to} $M$}
        \State $x_{t_i}^{*(r)} \leftarrow
               (1{-}t_i)\,x_0^{*(r)} + t_i\,\epsilon^{(r)}$
        \State Compute $v_i^{*(r)}$, $\Delta\mathbf{h}_{t_i}^{(r)}$,
               $\Delta\mathbf{e}_{t_i}^{(r)}$
    \EndFor
    \State Build $\mathcal{G}^{(r)}$ via Eqs.~\ref{eq:cache_rollout_short},\,\ref{eq:cache_deviation}
    \State $\bar{w} \mathrel{+}= \tfrac{1}{R}\,w^{(r)}$
\EndFor
\State \textbf{// Phase 2: Mean Graph (Eq.~\ref{eq:mean_edge})}
\State \textbf{// Phase 3: Shortest Path on Mean Graph (Eq.~\ref{eq:dp})}
\State $\mathcal{P}^{*}\leftarrow\texttt{ShortestPath\_DP}(\bar{\mathcal{G}},s,e,B{+}2)$
\State \textbf{// Phase 4: Inference along $\mathcal{P}^*$}
\State $\hat{x}_0\leftarrow\texttt{CachedDenoise}(\mathbf{c},\mathcal{P}^{*},
       \{\Delta\mathbf{h}_{t_i},\Delta\mathbf{e}_{t_i}\})$
\State \Return $\mathcal{P}^{*},\hat{x}_0$
\end{algorithmic}
\end{algorithm}

\noindent \textbf{Cost analysis.}
The dominant computational cost of IdealCache is the transformer forward passes. In the offline stage, we perform $R \cdot M$ full forward evaluations to construct reference trajectories, which constitute the main overhead.

In contrast, all subsequent operations for graph construction and scheduling are lightweight. Specifically, Phase~1 additionally performs $O(M^2)$ cached rollouts per reference run to estimate transition edges, where cached states reuse intermediate residuals and are significantly cheaper than full transformer evaluations. Phase~3 solves a dynamic program with complexity $O(M^2 B)$, which is negligible compared to even a single forward pass.

Importantly, the offline cost is incurred only once to build the mean graph $\bar{\mathcal{G}}$, and is amortized over all subsequent generations. At inference time, IdealCache executes exactly $B$ full forward passes and reuses cached computations for the remaining steps, resulting in the same number of full evaluations as standard caching-based methods under the same budget, without introducing additional runtime overhead.

\section{Proofs of Propositions}
\label{app:proofs}

\noindent \textbf{Proof of Proposition~\ref{prop:marginal}.}
Write $y = x_k^{*}$ and $z = x_t^{*}$. For any deterministic map $f$,
\begin{equation*}
    \mathbb{E}\,\|f(z) - y\|^2
    = \mathbb{E}\,\|f(z) - \mathbb{E}[y \mid z]\|^2
    + \mathbb{E}\,\|y - \mathbb{E}[y \mid z]\|^2
    + 2\,\mathbb{E}\,\bigl\langle f(z) - \mathbb{E}[y \mid z],\,
      \mathbb{E}[y \mid z] - y \bigr\rangle .
\end{equation*}
Conditioning on $z$, the cross term vanishes by the tower property, and the second term equals $\mathbb{E}\,\mathrm{Var}(y \mid z)$ (the trace of the conditional covariance), which does not involve $f$ and hence does not depend on the schedule. From Eq.~\ref{eq:ideal_traj}, $x_k^{*} = x_t^{*} - (t-k)(\epsilon - x_0^{*})$, so
$\mathbb{E}[x_k^{*} \mid x_t^{*}] = x_t^{*} - (t-k)\,\mathbb{E}[\epsilon - x_0^{*} \mid x_t^{*}] = x_t^{*} - (t-k)\,v^{\star}$. \hfill$\square$

\noindent \textbf{Proof of Proposition~\ref{prop:additive}.}
The rollout starts at $\hat{x}_t = x_t^{*}$. If the cached velocity estimate equals $v_t$ at every Euler sub-step in $(k,t]$, the sub-steps sum to $t-k$ and give $\hat{x}_k = x_t^{*} - (t-k)\,v_t$. On the reference trajectory, $x_k^{*} = x_t^{*} - (t-k)\,v^{\star}$ with $v^{\star} = \epsilon - x_0^{*}$. Subtracting gives $\hat{x}_k - x_k^{*} = (t-k)(v^{\star} - v_t)$ and $w(k\to t) = (t-k)^2\|v^{\star} - v_t\|^2$. This quantity depends only on $v_t$, $v^{\star}$ and $t-k$, and not on any other interval of the path, so the cost of a path is the sum of independent edge costs and the recursion in Eq.~\ref{eq:dp} returns the exact minimum. \hfill$\square$

\section{Detailed Experimental Setup}
\label{app:exp_setup}

\noindent \textbf{Models and Tasks.}
We evaluate on three training-free settings. Image editing is our primary setting: we use Qwen-Image-Edit~\citep{wu2025qwenimage} on GEdit-Bench~\citep{liu2025step1x} and FlowEdit~\citep{kulikov2024flowedit}, where preserving the source image's content is itself a quality target. We additionally evaluate on Qwen-Image~\citep{wu2025qwenimage} for text-to-image and HunyuanVideo~\citep{kong2024hunyuanvideo} for text-to-video generation.

\noindent \textbf{Ideal Trajectories and Schedule Grid.}
All methods discretise inference on $50$ timesteps uniformly spaced over $[0,1]$; all reported speedups are relative to this $50$-step \emph{vanilla pipeline}. For image editing, the source image directly provides $x_0^{\star}$ (Eq.~\ref{eq:ideal_traj}), making offline graph construction essentially free. For text-to-image and text-to-video, we use the $50$-step vanilla output as a proxy for $x_0^{\star}$ and linearly interpolate against the same noise realisation. Mean graphs for text-to-image and image editing are built from $R{=}1000$ prompts randomly sampled from the Pick-a-Pic dataset~\citep{kirstain2023pickapic}; for text-to-video we use $R{=}50$. Edge weights and shortest-path search follow Sec.~\ref{sec:graph}.

\noindent \textbf{Baselines and Metrics.}
We compare against FORA~\citep{selvaraju2024fora}, ToCa~\citep{zou2024toca}, DuCa~\citep{zou2024duca}, TaylorSeer~\citep{liu2025taylorseer}, HyCa~\citep{zheng2025hyca}, DiCache~\citep{bu2025dicache}, and TeaCache~\citep{liu2024teacache}, plus $50\%$ and $20\%$ step-reduction baselines. We report latency, FLOPs, and speedup against the vanilla pipeline together with task-specific quality metrics: ImageReward~\citep{xu2023imagereward}, SSIM~\citep{wang2004ssim}, and LPIPS~\citep{zhang2018lpips} for image tasks; SC, PQ, and OS on GEdit-Bench; LPIPS and SSIM for video.

\noindent \textbf{Notation and Hardware.}
We follow the symbol $\mathcal{N}$ as it appears in the baselines' tables; in FORA~\citep{selvaraju2024fora} this is explicitly the \emph{cache interval} (one full pass every $\mathcal{N}$ steps, intermediate steps reuse), and TaylorSeer and HyCa inherit the same parameterisation. For IdealCache we instead report $\mathcal{B}$, the global \emph{step budget} --- the total number of full transformer passes the schedule may spend, with their placement chosen by the shortest-path search of Sec.~\ref{sec:graph}. All experiments use a single NVIDIA H800 80GB GPU.

\section{Candidate-grid density is decoupled from inference cost}
\label{app:grid_density}

The candidate-grid size $M$ --- the number of equally spaced
timepoints in $[0,1]$ used to construct the planning graph --- is a
property of the graph, not of the sampler. A schedule chosen on a
denser graph is snapped back onto the standard $50$-step uniform grid,
so the cached-inference pipeline is identical across $M$. The main
results in Tab.~\ref{tab:t2i} use $M{=}50$ for strict parity with
caching baselines, whose grid \emph{is} the sampler. Tab.~\ref{tab:grid_density}
ablates the effect of densifying the grid on Qwen-Image / DrawBench at
$1664{\times}928$ with $B{=}10$.

\begin{table}[h]
\centering
\caption{Candidate-grid density ablation on Qwen-Image / DrawBench
($1664{\times}928$, IdealCache, $B{=}10$). SSIM and LPIPS are computed
against the $50$-step vanilla pipeline output.}
\label{tab:grid_density}
\renewcommand{\arraystretch}{1.0}
\small
\begin{tabularx}{\linewidth}{l *{2}{>{\centering\arraybackslash}X}}
\toprule
\textbf{Candidate grid $M$} & \textbf{SSIM}$\uparrow$ & \textbf{LPIPS}$\downarrow$ \\
\midrule
$50$  & 0.75 & 0.33 \\
$100$ & \textbf{0.78} & \textbf{0.27} \\
\bottomrule
\end{tabularx}
\end{table}

Doubling $M$ improves SSIM by $0.03$ and reduces LPIPS by $0.06$ at no
inference-cost penalty. Caching baselines cannot densify their grid
without proportionally adding forward passes; for IdealCache, $M$ is a
search-side knob that trades quality for offline planning budget alone.

\section{Additional qualitative examples}
\label{app:gallery}

\begin{figure}[h]
\centering
\includegraphics[width=\linewidth]{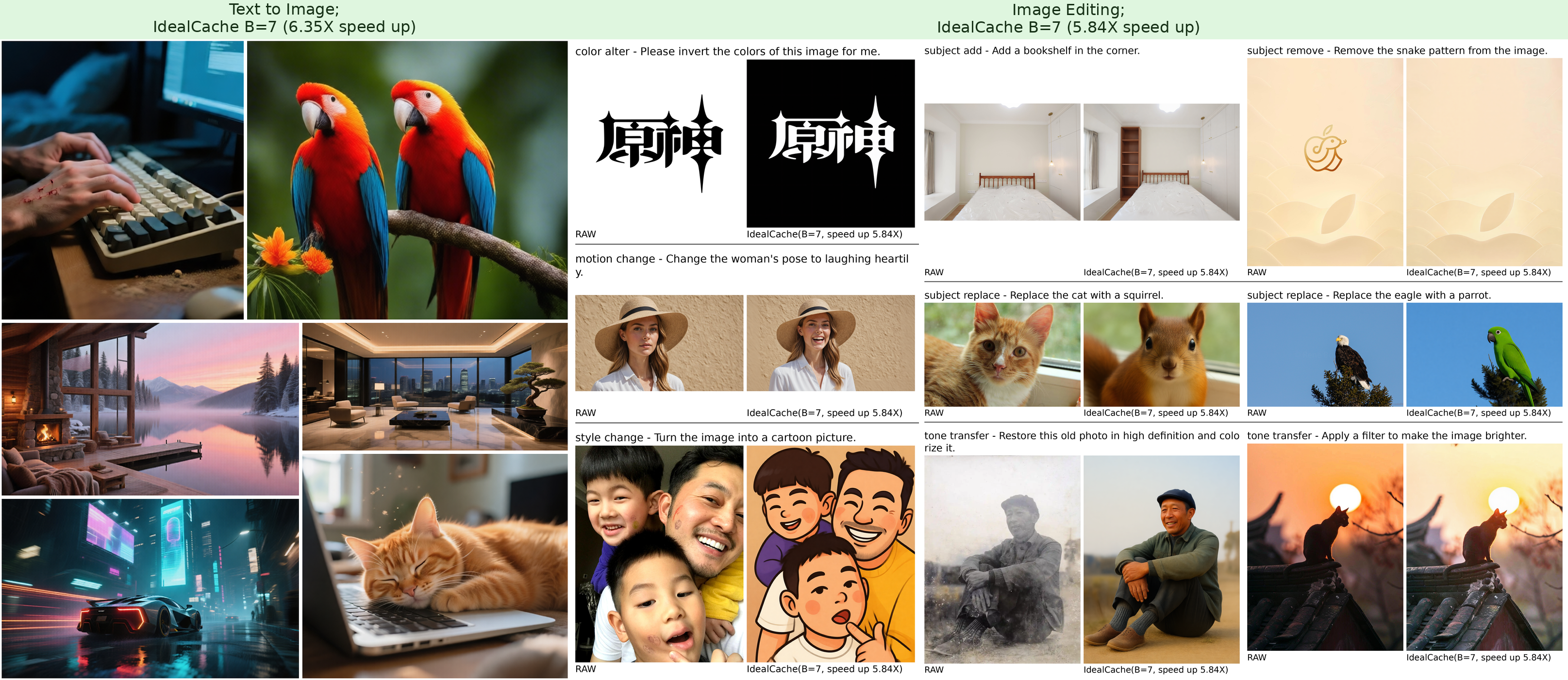}
\caption{Additional qualitative examples on Qwen-Image (text-to-image,
left) and Qwen-Image-Edit (image editing, right). IdealCache reaches
$6.35\times$ wall-clock speedup at $\mathcal{B}{=}7$ on text-to-image
generation and $5.84\times$ at $\mathcal{B}{=}7$ on image editing,
without retraining.The optimised
schedule preserves source-image content in non-edited regions while
faithfully applying the requested edit.}
\label{fig:gallery_t2i_edit}
\end{figure}

\begin{figure}[h]
\centering
\includegraphics[width=0.85\linewidth]{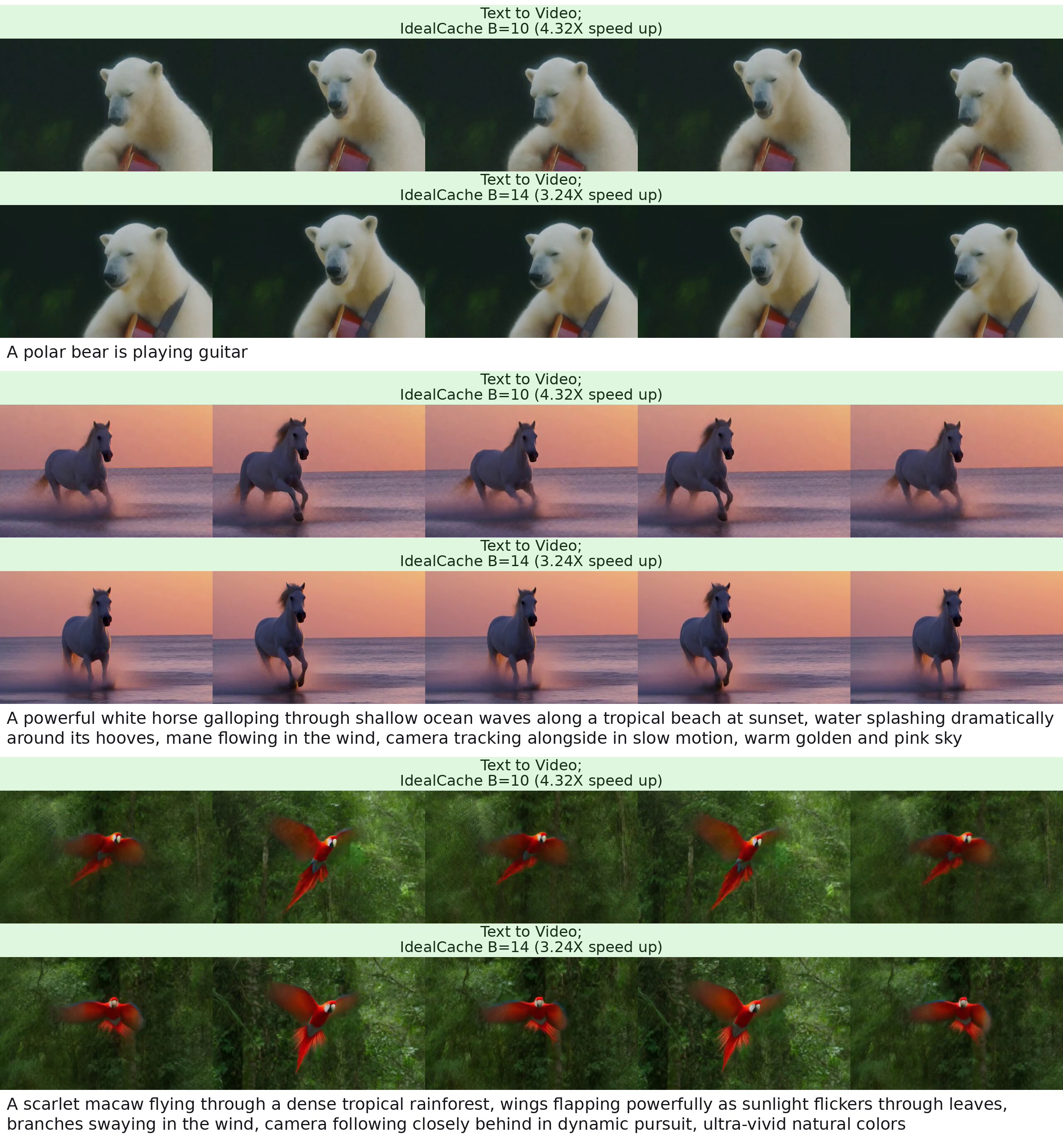}
\caption{Qualitative example on HunyuanVideo with IdealCache(153frames, 320x512 resolution).
Sampled frames from a generated clip; temporal coherence and visual
fidelity are preserved at substantially reduced compute relative to
the $50$-step vanilla pipeline.}
\label{fig:gallery_video1}
\end{figure}

\begin{figure}[h]
\centering
\includegraphics[width=0.85\linewidth]{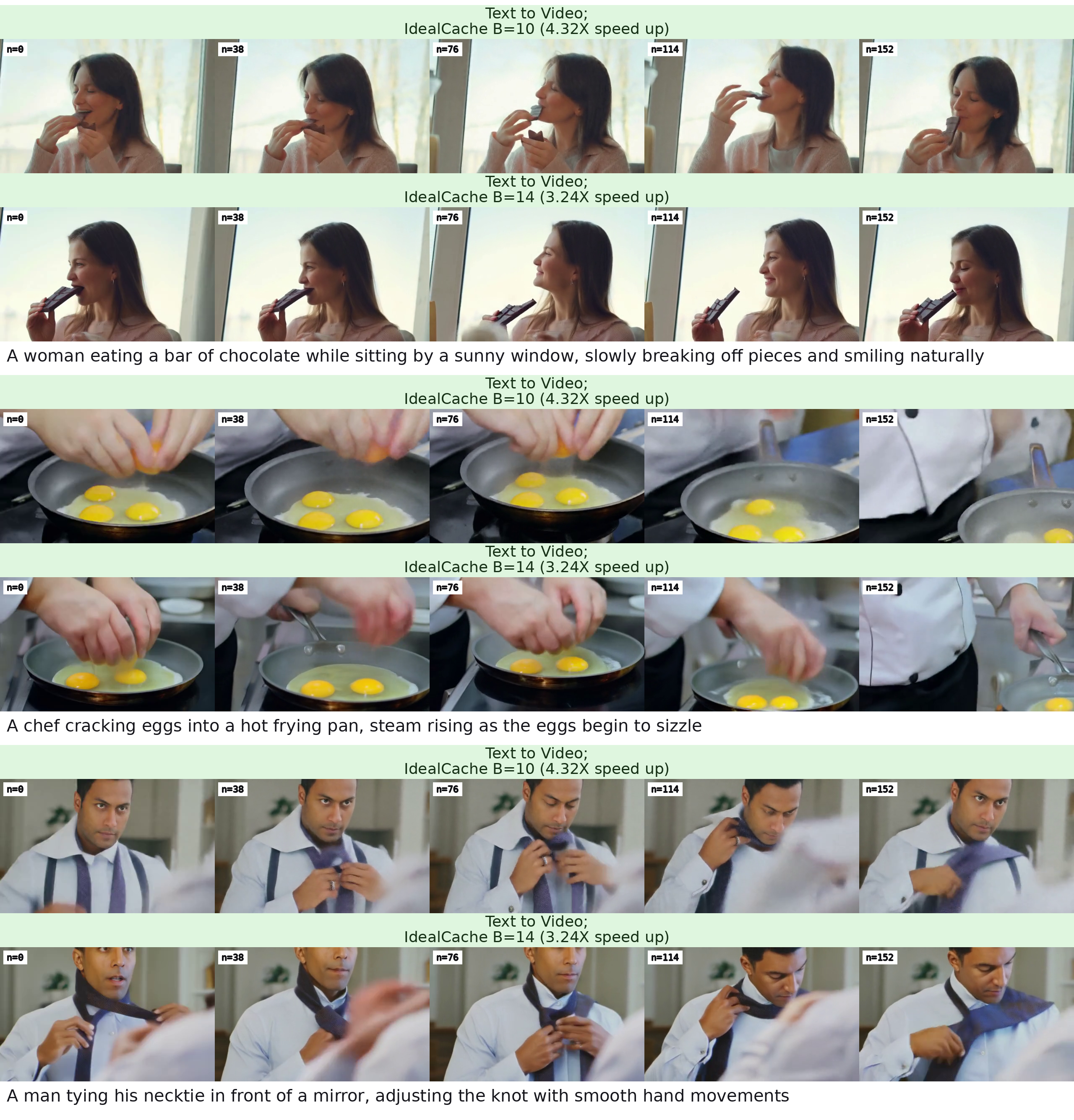}
\caption{Additional HunyuanVideo qualitative example under
IdealCache's accelerated schedule.}
\label{fig:gallery_video2}
\end{figure}

This appendix presents additional qualitative results across the three modalities. All examples are generated with IdealCache without retraining. The source models include Qwen-Image (text-to-image), Qwen-Image-Edit (image editing), and HunyuanVideo (text-to-video). Representative results are shown in Figures~\ref{fig:gallery_t2i_edit}, \ref{fig:gallery_video1}, and \ref{fig:gallery_video2}.

\section{Limitations}
\label{app:limitations}
Although IdealCache achieves state-of-the-art training-free acceleration and strong visual fidelity across multiple modalities, it has certain limitations. First, while we successfully reduce the offline graph-construction cost to near-zero for text-to-image tasks via one-shot sampling, tasks with high spatio-temporal heterogeneity (e.g., complex video generation) still require an ensemble of reference runs to construct a robust mean graph, meaning the offline cost is not universally negligible. Second, because our ideal trajectories for generation tasks are constructed using the vanilla pipeline's output as a proxy for the target (unlike in image editing, where the source image is used), our method is bounded by the generative capacity of the underlying Diffusion Transformer. If the base model fails to generate a coherent image or video at $50$ steps, IdealCache will optimally schedule a path toward that flawed target. Third, each task is evaluated on a single base model; validating IdealCache on additional models is left to future work. Finally, under extreme compression budgets (e.g., $\mathcal{B} < 5$), approximation errors from prolonged residual reuse inevitably compound, leading to some degradation in aggregate quality metrics. Nevertheless, because our framework is rooted purely in trajectory geometry and remains entirely architecture-agnostic, it can be seamlessly adapted to future, more powerful foundational models.

\section{Broader Impacts} % Or 'Social Impact' depending on conference template
The rapid scaling of Diffusion Transformers (DiTs) for high-resolution image and video synthesis has introduced prohibitive inference times and substantial computational costs. IdealCache directly addresses this bottleneck by providing up to a $6\times$ wall-clock speedup without requiring any model retraining, structural modifications, or complex distillation pipelines. This efficiency significantly democratizes access to state-of-the-art generative capabilities, allowing individual creators, small enterprises, and resource-constrained research labs to deploy heavy foundational models on accessible hardware. Furthermore, by drastically reducing the number of neural function evaluations (FLOPs) required per generation, our approach directly lowers the energy consumption and carbon footprint associated with generative AI inference at scale, contributing to greener and more sustainable AI ecosystems. To support the community and facilitate downstream research, we will make our code, graph-search algorithms, and evaluated schedules publicly available.

\section{LLM Usage Statement}
\label{sec:llm_usage}
In accordance with standard academic guidelines regarding the use of generative AI, the authors declare that Large Language Models (LLMs) were utilized solely for language polishing and stylistic refinement to improve the readability of this manuscript. All theoretical formulations, methodological designs, experimental executions, data analyses, and core intellectual contributions are entirely the original and independent work of the human authors.

\clearpage

\end{document}